\documentclass[10pt,twocolumn,letterpaper]{article}
\usepackage[pagenumbers]{iccv}

\usepackage{xspace}
\definecolor{cvprblue}{rgb}{0.21,0.49,0.74}
\usepackage[pagebackref,breaklinks,colorlinks,citecolor=cvprblue]{hyperref}
\usepackage{bbm}
\usepackage{duckuments}
\usepackage{multirow}
\usepackage{xfrac}

\usepackage{seqsplit}

\newcommand{\hash}[1]{{\ttfamily\seqsplit{#1}}}

\usepackage{amsmath,amsfonts,bm}

\def\eqref#1{equation~\ref{#1}}

\def\1{\bm{1}}

\newcommand{\bx}{\mathbf{x}}

\def\rvv{{\mathbf{v}}}

\def\rvx{{\mathbf{x}}}

\def\rmI{{\mathbf{I}}}

\def\vc{{\bm{c}}}

\DeclareMathAlphabet{\mathsfit}{\encodingdefault}{\sfdefault}{m}{sl}
\SetMathAlphabet{\mathsfit}{bold}{\encodingdefault}{\sfdefault}{bx}{n}

\newcommand{\bPi}{\boldsymbol{\Pi}}

\makeatletter
\renewcommand{\paragraph}{%
    \@startsection{paragraph}{4}%
    {\z@}{-0.5em}{-0.5em}%
    {\normalfont\normalsize\bfseries}%
}
\makeatother
\newcommand{\method}{Bolt3D\xspace}

\newcommand{\norm}[1]{\left\lVert#1\right\rVert}

\newcommand\lft{\mathopen{}\left}
\newcommand\rgt{\aftergroup\mathclose\aftergroup{\aftergroup}\right}

\newcommand{\mat}[1]{\mathbf{#1}}

\newcommand{\superscript}[1]{\mathit{#1}}

\def\paperID{2039}
\def\confName{ICCV}
\def\confYear{2025}

\usepackage{pifont}
\title{\method: Generating 3D Scenes in Seconds}

\author{
Stanislaw Szymanowicz$^{1,2}$
\quad
Jason Y. Zhang$^1$
\quad
Pratul Srinivasan$^{3}$ \\
\quad
Ruiqi Gao$^3$
\quad
Arthur Brussee$^3$
\quad
Aleksander Hołyński$^3$ \\ 
\quad
Ricardo Martin-Brualla$^1$
\quad
Jonathan T. Barron$^3$
\quad
Philipp Henzler$^1$
\vspace{2mm}
\\
\centerline{$^1$Google Research \quad $^2$VGG -- University of Oxford\quad $^3$Google DeepMind}
}

\begin{document}

\twocolumn[{%
\maketitle
\thispagestyle{empty}
\begin{center}
\centering
\captionsetup{type=figure}
\vspace{-1em}%
\includegraphics[width=\textwidth]{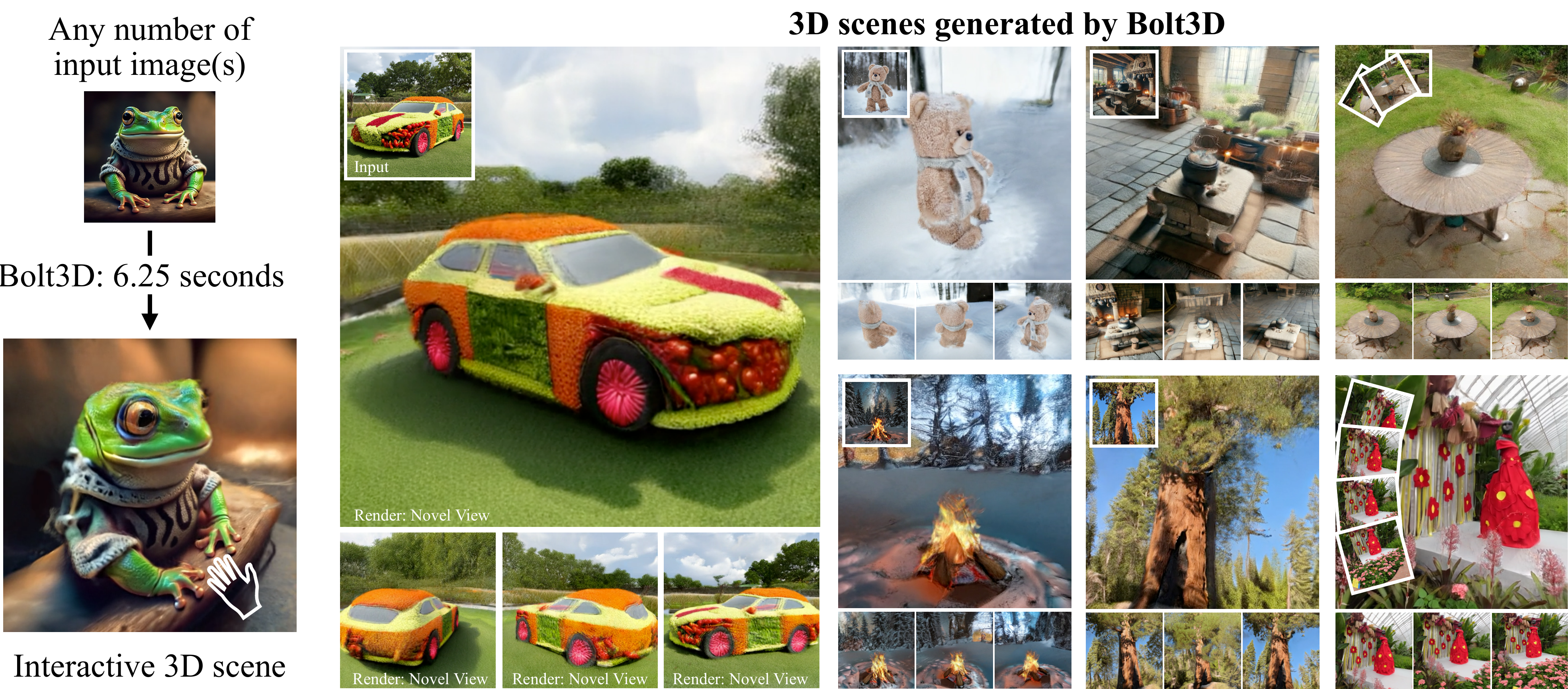}
\captionof{figure}{
Given an arbitrary number of input images, \method directly outputs a 3D representation which can be rendered at interactive frame-rates.
Operating in a feed-forward manner, generation takes mere seconds.
\method features a latent diffusion model with a scalable 2D architecture, trained on large-scale appearance and geometry data, enabling generation of full 360$^{\circ}$ scenes from one or multiple input images.
We invite the reader to explore these scenes in the interactive viewer available on the project website.
}%
\label{fig:teaser-fig}%
\end{center}%
\vspace{1em}%
}]

\begin{abstract}
\vspace{-0.7cm}

We present a latent diffusion model for fast feed-forward 3D scene generation. Given one or more images, our model \method directly samples a 3D scene representation in less than seven seconds on a single GPU.
We achieve this by leveraging powerful and scalable existing 2D diffusion network architectures to produce consistent high-fidelity 3D scene representations. To train this model, we create a large-scale multiview-consistent dataset of 3D geometry and appearance by applying state-of-the-art dense 3D reconstruction techniques to existing multiview image datasets. Compared to prior multiview generative models that require per-scene optimization for 3D reconstruction, \method reduces the inference cost by a factor of up to 300$\times$.
Project website: \texttt{\small \href{https://szymanowiczs.github.io/bolt3d}{szymanowiczs.github.io/bolt3d}}.

\end{abstract}
\section{Introduction}%
\label{sec:intro}

Modern image and video generative models generate compelling high-quality visual content, but these models sample 2D images, rather than an underlying 3D scene. 
The ability to directly generate 3D content instead would enable numerous applications, such as interactive visualization and editing.
However, scaling modern diffusion-based generative models to generate detailed 3D scenes remains a significant challenge for the research community, primarily due to two reasons.
First, representing and structuring (possibly unbounded) 3D data to enable training a diffusion model that generates full scenes at high resolution is an unsolved problem.
Second, ``ground truth'' 3D scenes are extremely scarce compared to the abundant 2D image and video data used to train state-of-the-art generative models.
As a result, many recent 3D generative models are limited to synthetic objects~\cite{lin2025diffsplat,xiang2024trellis,xudmv3d,zhang2024clay} or partial ``forward-facing'' scenes~\cite{liang2024wonderland,yu2024wonderjourney,yu2025wonderworld,wewer24latentsplat}.
Models that scale to real, full $360^{\circ}$ scenes use camera-conditioned multiview or video diffusion models to turn input image(s) into a large ``dataset'' of synthetic observations~\cite{gao2024cat3d,im3d}, from which an explicit 3D representation (such as a neural \cite{mildenhall2020nerf} or 3D Gaussian~\cite{kerbl3Dgaussians} radiance field) is then recovered via test-time optimization. While this approach is capable of producing high-quality 3D content, it is impractical; both sampling hundreds of augmented images with the multiview diffusion model and optimizing a 3D representation to match these images are slow and compute-intensive.

In this paper, we present a latent diffusion model for fast feed-forward 3D scene generation from one or more images. 
Our model, called \method, leverages the tremendous progress made in scalable 2D diffusion model architectures to generate an explicit 3D scene representation.
We represent 3D scenes as sets of 3D Gaussians, stored in multiple 2D grids in which each cell stores the parameters of one pixel-aligned Gaussian~\cite{szymanowicz24splatter,charatan23pixelsplat} (``Splatter Images''~\cite{szymanowicz24splatter}).
Crucially, unlike prior work, we use \emph{more Splatter Images} than input views, and we \emph{generate} Splatter Images using a diffusion model, which enables generating content for unobserved regions of the scene.

Our generation process consists of two parts: denoising color and position of each Gaussian and subsequently regressing each Gaussian's opacity and shape.
Given a set of posed input images and target camera poses, our model jointly predicts the scene appearance (pixel colors) viewed by the \emph{target} cameras as well as per-pixel 3D coordinates of scene points in \emph{all} cameras (both input and target).
We enable \method to predict accurate high-resolution per-pixel 3D geometry by designing and training a Geometry Variational Auto-Encoder (VAE~\cite{kingma14vae}) that resembles architectures used in image generation, but which we train from scratch using geometry data. 

Unlike 2D image datasets, real-world 3D scene datasets are small and limited.
To address this, we create a large-scale geometry dataset by running a robust Structure-from-Motion framework~\cite{mast3r} on large-scale multi-view image datasets. We use this new dataset to train our geometric VAE and diffusion model.

Finally, to obtain a renderable 3D representation (which 3D point coordinates alone do not directly provide), we train a Gaussian head network that takes the full set of high-resolution images, geometry maps, and camera poses predicted by the denoising network as input and predicts the remaining properties of 3D Gaussians (opacities, shapes), and refined colors.
The Gaussian head is supervised with rendering losses, which enables our full model's output to render high-quality novel views.

We demonstrate that \method  
outperforms prior single- and few-view feed-forward 3D regression methods and
synthesizes detailed 3D scene content, even in ambiguous regions that are not observed in any input image.
Furthermore, we show that \method reduces inference cost up to $300\times$ compared to multi-view image generation methods, which typically require per-scene optimization to reconstruct 3D models.
\begin{figure*}[t]
    \centering
     \includegraphics[width=\textwidth]{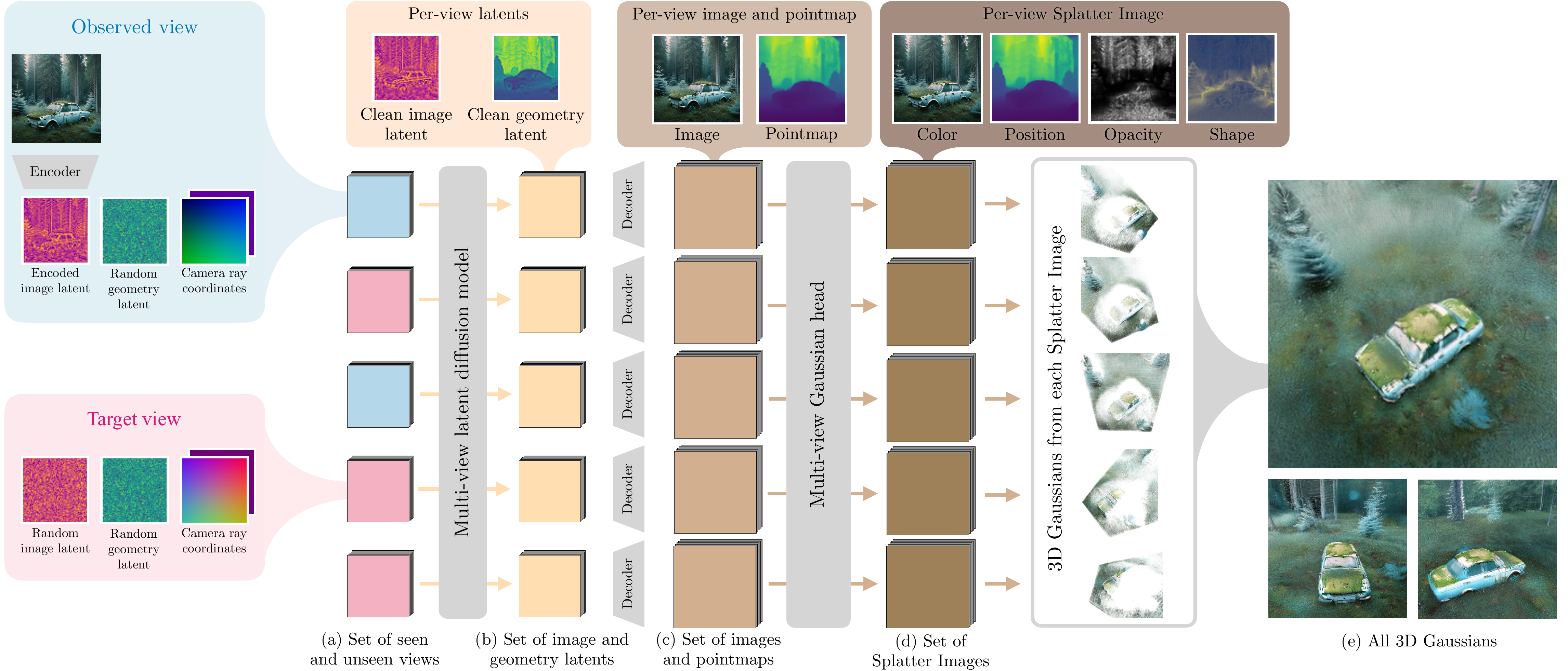}
    \caption{ {\bf Method.} 
    \method takes as input one or more posed, observed images, and a set of target poses (a), and outputs a renderable 3D scene (e).
    First, we use a multi-view latent diffusion model to sample per-view latent appearance and geometry (b).
    The appearance and geometry latents are independently decoded to full-resolution images and pointmaps (c) using a pre-trained image VAE decoder and our trained geometry decoder, respectively.
    Next, a multi-view Gaussian head predicts the opacities and scales of pixel-aligned 3D Gaussians, and refines the predicted colors.
    Together with the pointmap from  (c), these values form Splatter Images~\cite{szymanowicz24splatter} (d), which can be combined to create a complete 3D Gaussian representation of the scene (e).
     }
    \label{fig:method}
\end{figure*}

\section{Related work}%
\label{s:related}
 
\paragraph{Feed-forward 3D regression.}
Reconstructing a 3D scene from an input image (or from multiple images) is sometimes formulated as a feed-forward regression problem~\cite{tulsiani2018layer,li21mine,tatarchenko2017octree,tulsiani2017multi,groueix2018papier,gkioxari2019mesh,niemeyer2020dvr,chen2021mvsnerf,yu21pixelnerf,hong2024lrm,du2023learning,wu2023multiview,sajjadi2022srt,wang2021ibrnet}.
Most relevant to us are recent approaches that output 3D Gaussians~\cite{kerbl3Dgaussians}, a representation favored for its real-time rendering speed.
Splatter Image~\cite{szymanowicz24splatter} and pixelSplat~\cite{charatan23pixelsplat} were the first to associate Gaussians with the pixels of one or two input images, respectively.
Follow-up works iterated on this approach with improved architectures~\cite{gslrm2024,xu2024grm,tang2024lgm}, depth conditioning~\cite{szymanowicz2025flash3d,xu2025depthsplat}, explicit feature matching~\cite{chen2024mvsplat}, and removing camera pose requirements~\cite{ye2025noposplat,zhang2025FLARE}.
Although these regression-based methods are capable of accurately reconstructing observed regions, they tend to produce blurry results in unseen regions.
In contrast, \method is a generative approach and is thereby capable of generating unobserved regions of the scene.

\paragraph{Reconstruction via 2D image generation.}
A common paradigm for addressing ambiguity in few-view 3D reconstruction is to generate multiple views of the scene, from which 3D can be recovered.
Zero-1-to-3~\cite{liu2023zero1to3} and 3DiM~\cite{watson20223dim} did this for individual objects, and follow-up work improved the conditioning and sampling mechanism in a variety of ways~\cite{chan2023genvs, wu2024reconfusion, gu2023nerfdiff, hoellein2024viewdiff,zhou2022sparsefusion,sargent2023zeronvs,yu2023PhotoconsistentNVS}.
Joint sampling improves sample consistency, and can be achieved by multi-view  generators~\cite{gao2024cat3d,hoellein2024viewdiff,wallingford2025image,hu2024mvd,liu2023syncdreamer,shi2023MVDream,zhou2025stable}, or video models~\cite{chen2025mvsplat360,he2025cameractrl,im3d, voleti2024sv3d,liu2024reconx,sun2025dimensionx}. Alternatively, other methods have explored using estimated depth to improve geometric consistency~\cite{ma2025yousee,hu2024mvd,yu2024viewcrafter,yu2024wonderjourney,yu2025wonderworld,mueller2024multidiff}
The outputs of the generation process are 2D images, and reconstructing a renderable 3D asset requires an optimization-based reconstruction pipeline~\cite{mildenhall2020nerf, kerbl3Dgaussians}.
This two-stage approach produces high-quality results, but requires minutes or hours of optimization time for real-world scenes.
Our approach leverages a diffusion model to generate multiple views of the scene, but in our approach the 3D scene is a direct output from our model, so no optimization stage is required and the cost of inference is therefore decreased by multiple orders of magnitude.

\paragraph{3D Generation.}
Directly generating 3D representations circumvents the limitations of the previous works, by combining the ability of generative models to handle ambiguity while avoiding a costly distillation/optimization stage.
Early diffusion-based models denoise a voxel grid that parameterizes a radiance field~\cite{muller2023diffrf} and require 3D supervision. The difficulty in obtaining such 3D data motivated the formulation of a denoising objective on 2D images with a rendering bottleneck~\cite{anciukevicius2022renderdiffusion,anciukevicius2024denoising,szymanowicz2023viewset_diffusion,tewari2023diffusion}, similar to early GAN~\cite{goodfellow2014gan}-based works that directly output voxel grids~\cite{henzler2019platonicgan}, radiance fields~\cite{schwarz2020graf, yu2022generating,niemeyer2020giraffe} or tri-planes~\cite{bautista2022gaudi,chan2021eg3d} using differentiable rendering to directly train on 2D images. Follow-up works improve the architecture~\cite{xudmv3d} and the 3D representation~\cite{wang2024splatflow, lin2025diffsplat}, but such approaches have shown limited success beyond individual objects and small baseline camera motion. Some more recent works~\cite{liang2024wonderland,schwarz2025ggs} exploit temporal locality to generate videos given camera trajectories. 
A second body of work aims to directly learn a `natively 3D' latent space, showing remarkable success on bounded objects~\cite{lin2025diffsplat,meng2024zero1tog,roessle2024l3dg,xiang2024trellis,zhang2024clay,zhang20233dshape2vecset}, but lagging behind in quality when applied to 3D scenes with backgrounds~\cite{kim2023neuralfieldldm,ren2024scube,schwarz2024wildfusion,yang2024prometheus} due to limitations of the autoencoder.
We opt for a view-centric representation and analyze the architecture and training recipe of the autoencoder to achieve very high visual fidelity.
Most similar to our work is latentSplat~\cite{wewer24latentsplat} which learns a latent geometry representation similar to ours. However, latentSplat is limited to single object categories, is only applicable to 2-view reconstruction and features a VAE-GAN framework.
Our model works on any object or scene, can take any number of images as input and builds on a powerful latent diffusion model. We also demonstrate that our method performs better.

\section{Preliminaries}
\label{s:preliminaries}

\newcommand{\x}{\boldsymbol{x}}
\newcommand{\y}{\boldsymbol{y}}
\newcommand{\bbb}{\boldsymbol{b}}
\newcommand{\bu}{\boldsymbol{u}}
\newcommand{\bc}{\boldsymbol{c}}
\newcommand{\bv}{\boldsymbol{v}}
\newcommand{\bq}{\boldsymbol{q}}
\newcommand{\bs}{\boldsymbol{s}}
\newcommand{\bG}{\boldsymbol{G}}
\newcommand{\bnu}{\boldsymbol{\nu}}
\newcommand{\balpha}{\boldsymbol{\alpha}}
\newcommand{\bmu}{\boldsymbol{\mu}}
\newcommand{\bsigma}{\boldsymbol{\sigma}}
\newcommand{\super}[1]{\text{#1}}
\newcommand{\I}{\mathrm{I}}

\paragraph{Latent Diffusion Models.} The key component of our method is a latent diffusion model~\cite{ho2020denoising,song2020denoising}. A diffusion model defines a forward process by gradually adding Gaussian noise to a data point $\rvx_0$: $p(\rvx_t | \rvx_0) = \mathcal{N}(\alpha_t \rvx_0, \sigma_t)$, until $\rvx_t$ is close to a Gaussian. A denoiser model $\hat{\rvx}_\theta(\rvx_t; t)$ is learned to predict the clean sample $\rvx_0$ given the noisy sample $\rvx_t$ at time step $t$ by minimizing a weighted $\ell_2$ loss:
\begin{equation}
    \theta = \operatorname{arg\,min}_{\theta} \mathbb{E}_{p_{\rvx_0},p(\rvx_t|\rvx_0) } w(t) \norm{\hat{\rvx}_\theta(\rvx_t; t) - \rvx_0 }^2_2
\end{equation}
Other parametrizations of the denoiser model have been proposed, such as predicting the noise added to the clean data or a combination of the clean data and the noise like the $\rvv$-prediction~\cite{salimans2022progressive}.
A latent diffusion model first compresses the data into a lower-dimensional latent space and then builds a diffusion model in the latent space. A two-stage training procedure is typically applied: the compression is first learned by a VAE and fixed, followed by learning the diffusion model.

\paragraph{3D Gaussian Representation.} We leverage a set of 3D Gaussians as the scene representation: $\mathcal{G} = \{ \bmu_i, \sigma_i, \boldsymbol{\Sigma}_i, \vc_i \}_{i=1}^G $ where $\bmu_i \in \mathbb{R}^3$ is the mean of the Gaussian, $\sigma_i \in \left[ 0, 1 \right]$ its opacity, $\boldsymbol{\Sigma}_i \in \mathbb{R}^{3 \times 3}$ its covariance matrix and $\vc_i \in \mathbb{R}^3$ is the isotropic color. These Gaussians can be rendered to a camera efficiently using the splatting operation~\cite{kerbl3Dgaussians}.

\paragraph{Few-view 3D reconstruction.} The goal of few view reconstruction is to recover the representation $\mathcal{G}$ given a small collection of $N$ input views, where ``views" are paired images $\rmI$ and their camera poses $\bPi$. In our case, we assume $N \in \{1-4\}$. With such limited input views and therefore insufficient coverage of the scene, pure optimization-based reconstruction becomes an ill-posed problem, as many representations $\mathcal{G}$ can explain the same set of posed images $(\mat{I}, \bPi)$. Therefore, we instead formulate the task as a generation problem: learn and sample from a conditional distribution of the 3D scene given the input views: $p(\mathcal{G}|\rmI, \bPi)$.

\section{Method}%
\label{s:method}

\method takes a single or multiple images and their camera poses as input, and outputs a 3D Gaussian representation. The model consists of two components: a latent diffusion model that generates more views and the 3D location per pixel (i.e., a 3D pointmap) for each view, and a feed-forward Gaussian head model that takes the outputs from the diffusion model and predicts full parameters of one colored 3D Gaussian per pixel (i.e. a Splatter Image) for each view. Below we describe the 3D Gaussian representation (\cref{s:representation}), the latent diffusion model (\cref{s:ldm}), the feed-forward Gaussian head model (\cref{s:gaussian-head}) and the training process (\cref{s:training}).
We present an overview in~\cref{fig:method}.

\subsection{3D Representation}\label{s:representation} 
Assuming $K$ views that can be observed or generated, we leverage a 3D Gaussian representation consisting of $K$ Splatter Images~\cite{szymanowicz24splatter}, i.e., pixel-aligned colored 3D Gaussians for each view. Each 3D Gaussian contains four properties: color, 3D position (mean of the Gaussian), opacity, and covariance matrix.
In contrast to prior reconstruction-based methods, we leverage a generative approach that hallucinates new content in a wider range of the scene than what is covered by the input views.

\paragraph{Factorized sampling.} 
We factorize the generation of Gaussian parameters into two parts: first, a latent diffusion model is leveraged to generate the color and 3D position. Then, a feed-forward Gaussian head model takes the color and 3D position as input and predicts the opacity, covariance matrix and refined color. The motivation is that we can easily get data of the colors and 3D positions by collecting captured images and running dense Structure-from-Motion to serve as the target of the latent diffusion model. On the other hand, finding direct supervision for the covariance matrices and opacities is non-trivial. However, given the colors and 3D positions generated from the diffusion model, the covariance matrix and opacity are much less ambiguous, and therefore can be modeled by a deterministic mapping function that is supervised with a rendering loss.

\begin{figure*}[t]
    \centering
    \includegraphics[width=\linewidth]{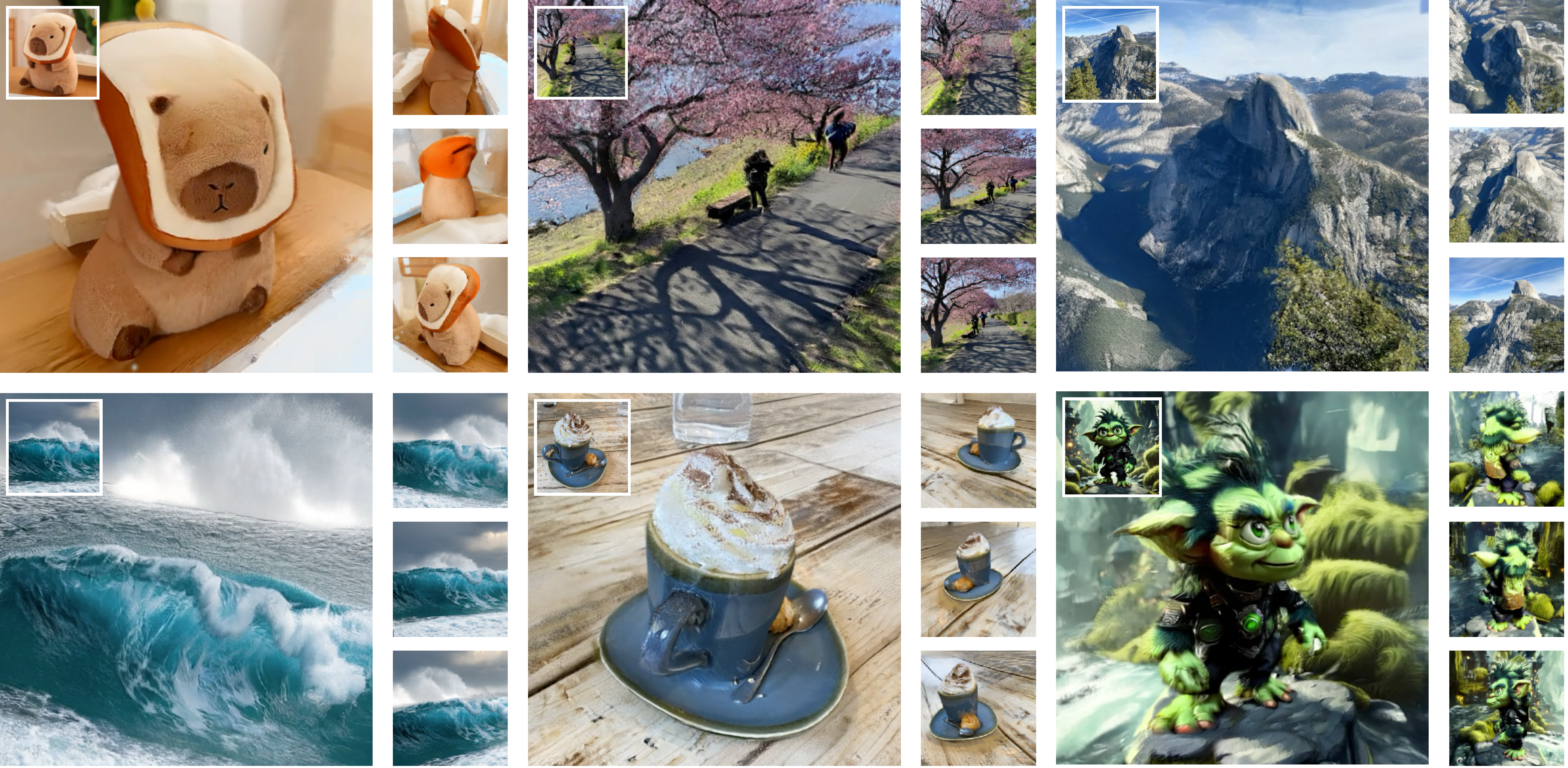}
    \caption{\textbf{Qualitative results.} We show renders of our 3D scenes reconstructed from just one input image (top left corner in each image) in a feed-forward manner.
    Inference takes only 7 seconds on a single GPU.}
    \label{fig:qualitative_results}
\end{figure*}

\subsection{Geometric multi-view latent diffusion model} \label{s:ldm}
We train a multi-view latent diffusion model that jointly models images and 3D pointmaps. Specifically, the model takes one or more images $\mat{I}^{\rm cond}$ and their camera poses $\mat{\Pi}^{\rm cond}$ as input. Given multiple target camera poses $\mat{\Pi}^{\rm tgt}$, the model learns to capture the joint distribution of the target images $\mat{I}^{\rm tgt}$, the target pointmaps $\mat{P}^{\rm tgt}$ and the source view pointmaps $\mat{P}^{\rm cond}$:

\begin{equation}
p(\mat{I}^{\rm tgt}, \mat{P}^{\rm tgt}, \mat{P}^{\rm cond}| \mat{I}^{\rm cond}, \mat{\Pi}^{\rm cond}, \mat{\Pi}^{\rm tgt})
\end{equation}

\paragraph{Model architecture.} We finetuned our model from a pretrained multi-view image diffusion model, which itself was finetuned from text-to-image latent diffusion model, to maximally maintain the generalization ability of the model while being trained on limited amount of multi-view data with 3D positions. The images and geometry (i.e., pointmaps and cameras) are encoded and decoded by two separate VAEs that $8\times$ downsample the input signals spatially (i.e. from $512 \times 512$ in the pixel space to $64 \times 64$ in the latent space). The image VAE is pre-trained and frozen while the geometry VAE is trained from scratch. The camera pose is parametrized as a $6$-dim raymap that encodes the ray origin and direction at each spatial location. At the input of the latent diffusion, the image latent, geometry latent and a raymap of the same size of the latents are channel-wise concatenated. We use a $\rvv$-parametrization and a $\rvv$-prediction loss for the diffusion model~\citep{salimans2022progressive}. The model is pre-trained on $8$ and fine-tuned on 16 views in total, and $1$-$3$ views are randomly sampled as input views. During sampling we generate $16$ views in total.

\paragraph{Geometry VAE.}
\label{s:vae}

We train a geometry VAE to jointly encode the pointmap $\mat{P}$ and camera raymap $\mat{r}$ of a view into a geometry latent: 
\small
\begin{equation}
\bmu, \bsigma = \mathcal{E}(\mat{P}, \mat{r})\,, \,\,\, \mathbf{z} \sim \mathcal{N}(\bmu, \bsigma)\,, \,\,\, \hat{\mat{P}}, \hat{\mat{r}} = \mathcal{D}(\mathbf{z})\,,
\end{equation}
\normalsize
where $\mathcal{E}$ is a convolutional encoder and $\mathcal{D}$ is a transformer decoder. The model is optimized by minimizing the following training objective, which is a combination of the standard VAE objective and a geometry-specific loss:
\begin{equation}
\mathcal{L}= \mathcal{L}_{\text{rec}} + \lambda_{1}\mathcal{L}_{\text{KL}} + \lambda_2\mathcal{L}_{\text{grad}}.
\end{equation}
The reconstruction loss $\mathcal{L}_{\text{rec}}$ is given by 
\begin{equation}
    \mathcal{L}_{\text{rec}} = \| \mat{w}(\hat{\mat{P}} - \mat{P}) \|_2^2 + \|\hat{\mat{r}}- \mat{r}\|_{2}^{2},
    \label{eq:reconstruction_loss}
\end{equation}
where $\mat{w}$ is a per-pixel weighting depending on the distance from the point to the center of the scene in the local camera frame (see supplement for more details). Intuitively, this encourages accurate geometry while accounting for lower confidence further from the camera. $\mathcal{L}_{\text{KL}}$ is  given by: 
\begin{equation}
\mathcal{L}_{\text{KL}} = -\frac{1}{2}\sum_k \lft(1 + \log \sigma_{k} - \mu_{k}^2 - \sigma_k \rgt)\,.
\end{equation}
Finally, we add an $\ell_2$ reconstruction error of the vertical and horizontal gradients of the pointmap, to improve boundary sharpness in the decoded pointmap:
\begin{equation}\label{eq:grad_loss}
\mathcal{L}_{\text{grad}} = \|\partial_u\hat{\mat{P}} - \partial_u \mat{P}\|_2^2 + \|\partial_v\hat{\mat{P}} - \partial_v \mat{P}\|_2^2\,.
\end{equation}

\subsection{Gaussian Head} \label{s:gaussian-head}
Given cameras and generated images and pointmaps, we train a multi-view feedforward Gaussian head model to output the refined color, opacities and covariance matrices of 3D Gaussians stored in Splatter Images. We first calibrate the generated pointmaps to be pixel aligned under the corresponding camera coordinate system. That is, we transform the 3D points into the camera coordinate, keep $z$ and set $x, y$ based on the camera ray and $z$. While the VAE decodes color and pointmaps independently for each view, we found that implementing a multi-view Gaussian head model is crucial. A U-ViT~\cite{bao2022all} architecture is applied with $4\times$ patchification before the transformer blocks, see Supplementary for more details.
The Gaussian head takes 8 views as input, and is trained using photometric losses (i.e. an L2 loss and a perceptual loss~\cite{zhang2018perceptual}) on rendered images from $4$ subsampled input views and $8$ novel views.

\subsection{Training}
\label{s:training}
\paragraph{Data.}
Our method requires supervision of dense, multi-view consistent pointmaps associated with the multi-view images.
We leverage a recent state-of-the-art dense reconstruction and matching method, MASt3R~\cite{mast3r}, running the depth and feature estimation followed by bundle adjustment of all pixels for 20-25 images per scene.
This setup allows for complete scene coverage and outputs multi-view consistent 3D pointmaps.
While the data is not perfect, we find that the residual noise and geometric imperfections are minor enough to not affect our method significantly (partially due to using rendering losses, in addition to geometry losses).
We run MASt3R on all scenes from CO3D~\cite{reizenstein21co3d}, MVImg~\cite{yu2023mvimgnet}, RealEstate10K (RE10K)~\cite{zhou2018stereo} and DL3DV-7K~\cite{ling2024dl3dv} forming a dataset of around 300k multi-view consistent 3D scenes.
We use standard train-test splits in CO3D, MVImg and RE10K, and  we use the first 6K scenes in DL3DV for training and last 1K for testing.
In addition, we leverage synthetic object datasets (Objaverse~\cite{deitke2023objaverse} and a high-quality internal object dataset) with their corresponding pointmap renderings.
We train on a mix of these datasets, sampling the real scenes with equal probability from all 4 real datasets and sampling the synthetic datasets 1:2 compared to sampling real data.

\paragraph{Training Protocol.}
We train our model in 3 stages:
\begin{enumerate}
    \item Geometry VAE. We first train the model at 256$\times$256 resolution for 3 million iterations, then finetune at 512$\times$512 for another 250k iterations.
    \item Gaussian head. Given ground truth color and autoencoded geometry, the Gaussian head is trained for 100k iterations with rendering losses to output Splatter Images.
    \item Latent diffusion model. We initialized our latent diffusion model from CAT3D~\cite{gao2024cat3d} and trained it for 700k iterations on the 8-view setup before finetuning it for 70k iterations on 16-views.
    See supplement for more details.
\end{enumerate}
\section{Experiments}\label{s:experiments}

We begin our experiments with presenting the results of our method in~\cref{fig:qualitative_results}, and we encourage the reader to view the videos and interactive visualizations on the project website which include high-quality reconstructions on a wide range of inputs.

Our experiments are divided into 4 sections.
First, we illustrate that modeling ambiguity using a generative model is crucial for few-view reconstruction by evaluating against state-of-the-art regression-based approaches.
Second, we show that our approach for modeling ambiguity outperforms recent approaches for feed-forward 3D generation from few input images.
Third, we evaluate the speed-quality trade-off between state-of-the-art optimization-based methods and our method.
Finally, we analyze the Geometry VAE and show its crucial role in the performance of our method. For Geometry VAE and gaussian decoder ablations we refer to the supplement.

When comparing against prior works we evaluate our method on the number of views other methods were designed for.
We evaluate performance at center crops of $512\times512$ resolution unless stated otherwise.

\paragraph{Metrics.} We quantify 3D reconstruction quality with standard metrics for novel-view synthesis: PSNR, SSIM, LPIPS~\cite{zhang2018perceptual} and FID~\cite{heusel2018fid} to measure pixel-wise, patch-wise, perceptual and distribution similarity, respectively.

\subsection{Comparison to 3D Regression.}

We compare our method with state-of-the-art feed-forward Gaussian Splat regression methods: Flash3D~\cite{szymanowicz2025flash3d} (single-view) and DepthSplat~\cite{xu2025depthsplat} (few-view).

\paragraph{Protocol.} We use 1-view RE10K for comparison against Flash3D by using the split from~\cite{wu2024reconfusion,gao2024cat3d} and using the first frame in each video as the conditioning frame, with the target frames as specified in the split.
This split contains larger camera motion (90 frames) than that used in the original Flash3D paper, thus evaluating the reconstruction ability  beyond small ($\pm$ 30 frames) camera motion.
Additionally, we include an evaluation on CO3D, also adapted from~\cite{wu2024reconfusion,gao2024cat3d} 
to evaluate both methods under stronger self-occlusion, as is typical for single-view reconstruction.
For comparison against DepthSplat~\cite{xu2025depthsplat}, we use DL3DV, which features large camera motion, and we evaluate both methods in the 2-view and 4-view settings, using source and target views from~\cite{xu2025depthsplat}.
We run evaluations on the scenes using the overlap of their and our testing sets.

\begin{figure}
    \centering
    \includegraphics[width=0.99\columnwidth]{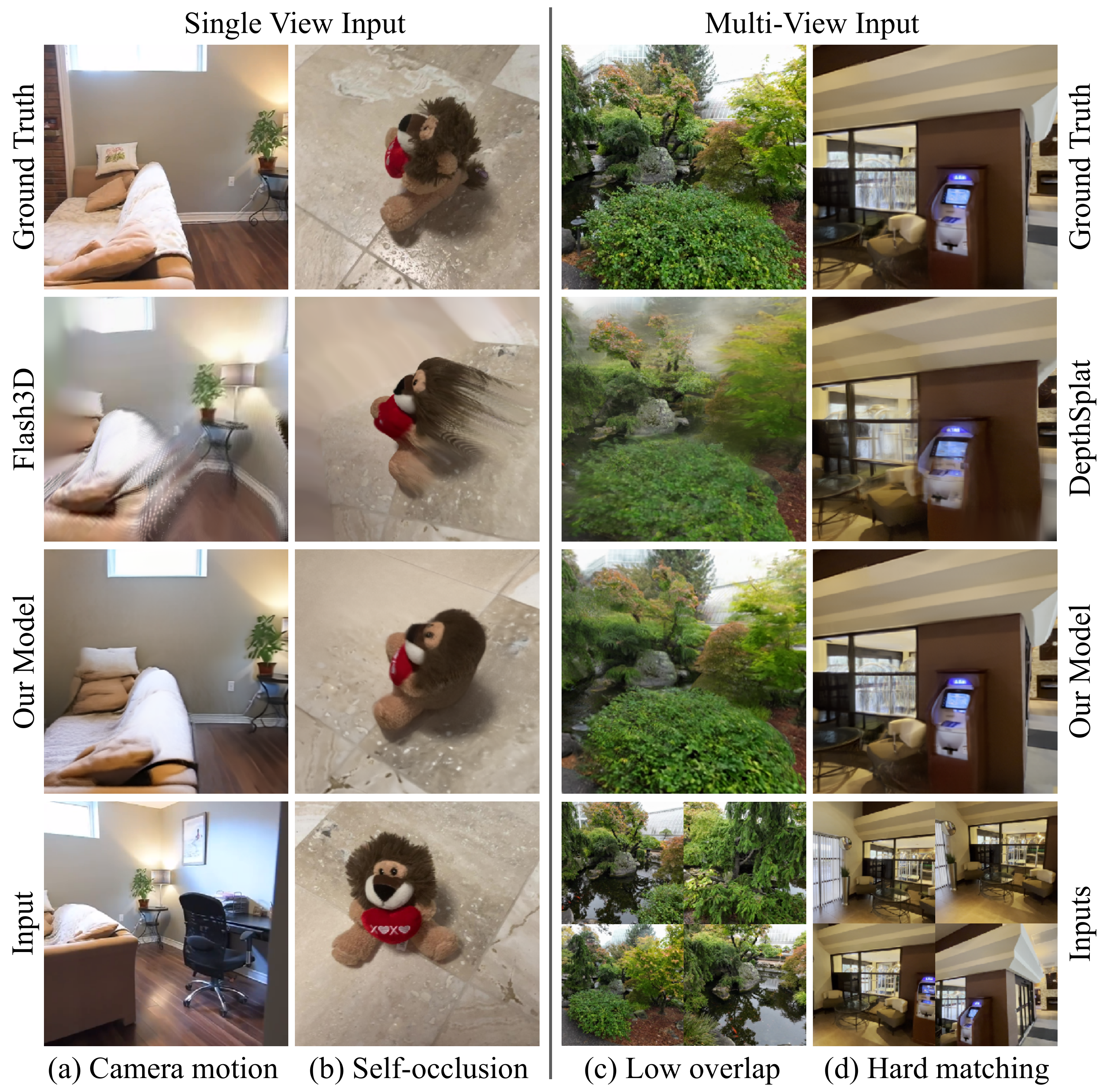}
    \vspace{-6pt}
    \caption{\textbf{Modeling ambiguity} is advantageous in both single- and few-view reconstruction. Given just a single input image, our generative framework  synthesizes realistic unobserved outside the observed field-of-view (a) and in occluded regions (b). \method also performs well in challenging scenarios given multiple input images that have a small overlap (c) or cases where feature matching is difficult (d).}
    \label{fig:modeling_ambiguity}
\end{figure}

\paragraph{Results.} We find that \method outperforms both methods, as shown in \cref{tab:regressive_reconstruction}.
The strong performance of our method compared to feed-forward reconstruction highlights the importance of modeling ambiguity, as we show in \cref{fig:modeling_ambiguity}.
This observation is consistent with the fact that the biggest gain in performance is observed in the 1-view setting, where ambiguity is the largest.
        
\begin{table}
    \centering
    \footnotesize
    \resizebox{0.97\columnwidth}{!}{%
    \begin{tabular}{llccccc}
    \toprule
        {} & Method    & PSNR $\uparrow$ & SSIM $\uparrow$ & LPIPS $\downarrow$ & FID $\downarrow$ \\
        \midrule
        1-view  & Flash3D  & 17.40 & 0.699 & 0.419 & 96.9 \\
        RE10K   & Ours  & \textbf{21.03} & \textbf{0.805} & \textbf{0.257} & \textbf{55.5} \\
        \midrule 
        1-view    & Flash3D & 14.43 & 0.552 & 0.608 & 174.8 \\
        CO3D      & Ours & \textbf{16.78} & \textbf{0.562} & \textbf{0.505} & 
        \textbf{97.5} \\
        \midrule
        2-view   & DepthSplat & 16.25 & 0.515 & 0.465 & 95.9 \\
        DL3DV    & Ours & \textbf{17.75} & \textbf{0.551} & \textbf{0.392} & 
        \textbf{64.5}
        \\
        \midrule
        4-view   & DepthSplat & 19.48 & 0.638 & 0.327 & 58.8 \\
        DL3DV    & Ours & \textbf{20.64} & \textbf{0.653} & \textbf{0.310} &
        \textbf{48.2} 
        \\
         \bottomrule
    \end{tabular}
    }
    \caption{\textbf{Comparison to regression-based methods.}
    Our generative approach improves performance across multiple datasets and numbers of input views. The biggest gain is seen in the 1-view setting where ambiguity is the greatest.}
    \label{tab:regressive_reconstruction}
\end{table}

\subsection{Comparison to Feed-Forward 3D Generation.}\label{s:comparison_generation}

We compare our method to two recent feed-forward 3D generative methods capable of reconstructing real scenes: LatentSplat~\cite{wewer24latentsplat} and Wonderland~\cite{liang2024wonderland}.

\paragraph{Protocol.}
When comparing to LatentSplat, we evaluate at 256$\times$256 resolution on the datasets they proposed: RealEstate10k and the hydrants and teddybear categories of CO3D, with their extrapolation splits.
When comparing to Wonderland, we follow their quantitative evaluation protocol on RealEstate10K, i.e. randomly sampling 1000 testing scenes, using the first frame of the video as the source frame and measuring the rendering quality of 14 views following the source view given that one input view.

\begin{figure}
    \centering
    \includegraphics[width=0.99\columnwidth]{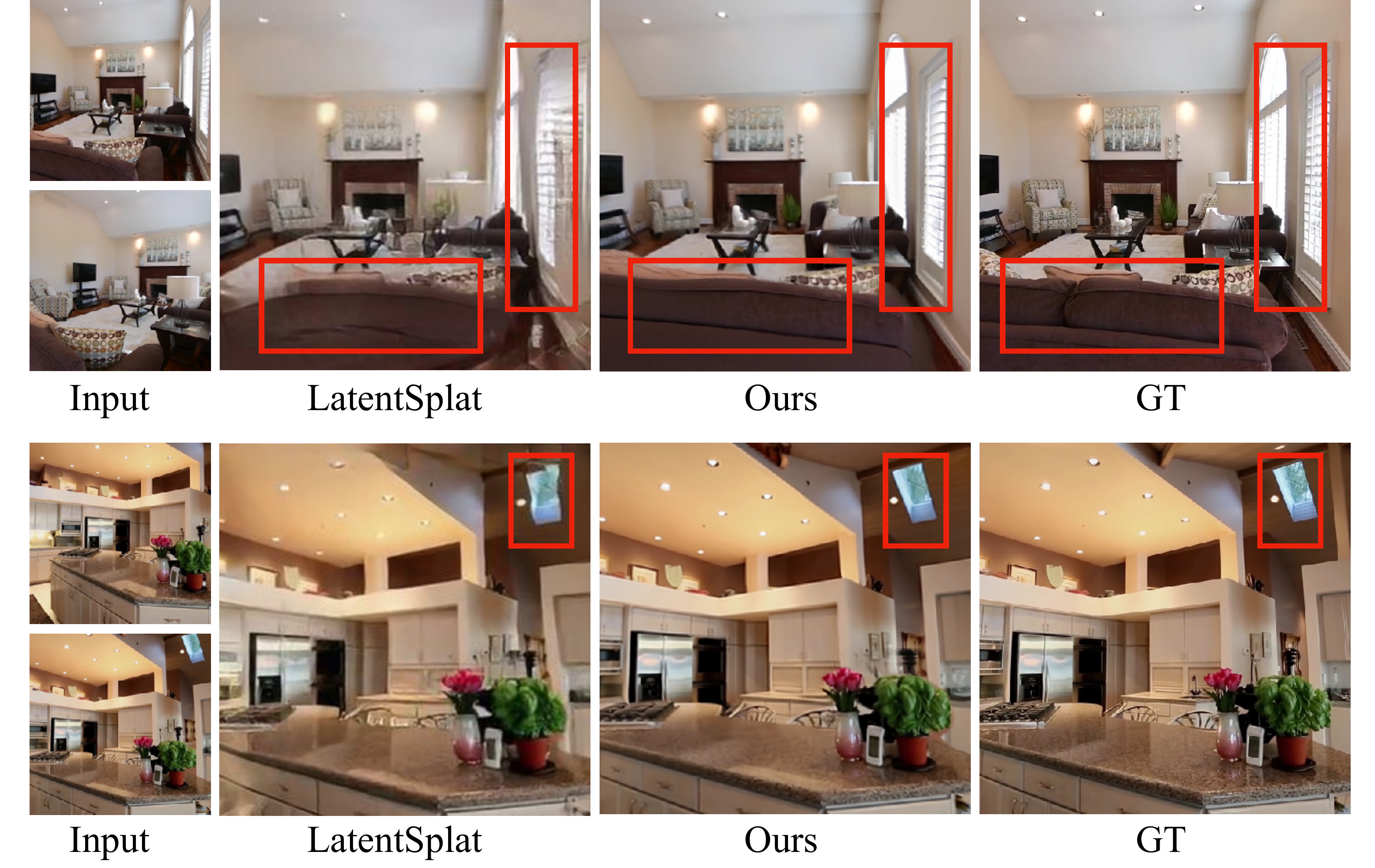}
    \vspace{-6pt}
    \caption{Our latent diffusion framework generates more realistic details the VAE-GAN framework proposed by LatentSplat.}
    \label{fig:comparison_latentsplat}
\end{figure}

\paragraph{Results.}
We observe that our method outperforms LatentSplat qualitatively (\cref{fig:comparison_latentsplat}) and quantitatively (\cref{tab:generative_reconstruction}) on image-level metrics.
We suspect that latentsplat's adversarial loss gives it an advantage in the distribution-level FID metric.
\cref{fig:comparison_latentsplat} illustrates that our diffusion model generates higher-quality details than LatentSplat's VAE-GAN.
We also demonstrate that our method outperforms Wonderland in \cref{tab:generative_reconstruction}.
We hypothesize this is due to explicit modeling of geometry in the autoencoder and in the diffusion model.
It is also worth noting that Wonderland uses a video model and takes 5 minutes to generate a scene, while ours generates 16 splatter images in around 6 seconds.

\begin{table}
    \centering
    \footnotesize
    \resizebox{0.97\columnwidth}{!}{%
    \begin{tabular}{llcccc}
    \toprule
        {} & Method    & PSNR $\uparrow$ & SSIM $\uparrow$ & LPIPS $\downarrow$ & FID $\downarrow$ \\
        \midrule
        1-view  & Wonderland* & 17.15 & 0.550 & 0.292 & - \\
        Re10k   & Ours & \textbf{21.54} & \textbf{0.747} & \textbf{0.234} & 11.30 \\
        \midrule
        2-view  & LatentSplat  & 22.62 & 0.777 & 0.196 & \textbf{2.79} \\
        RE10K   & Ours & \textbf{23.13} & \textbf{0.806} & \textbf{0.166} & 3.66 \\
        \midrule
        2-view & LatentSplat  & 17.71 & 0.533 & 0.434 & \textbf{71.12} \\
        CO3D ted. & Ours & \textbf{18.94} & \textbf{0.605} & \textbf{0.393} & 76.24 \\
        \midrule
        2-view    & LatentSplat  & 15.78 & 0.306 & 0.426 & \textbf{48.03} \\
        CO3D hyd. & Ours & \textbf{17.38} & \textbf{0.437} & \textbf{0.390} & 76.12 \\
         \bottomrule
    \end{tabular}
    }
    \caption{\textbf{Comparison to 3D generative methods.} Our diffusion-based framework with explicit 3D supervision achieves the best image-level metrics. * denotes results reported in~\cite{liang2024wonderland}}
    \label{tab:generative_reconstruction}
\end{table}

\subsection{Optimization-based 3D Reconstruction.}

We explore the speed-quality trade-off versus a state-of-the-art optimization-based method, CAT3D~\cite{gao2024cat3d}.
We also include qualitative comparisons to RealmDreamer~\cite{shriram2024realmdreamer}.

\paragraph{Protocol.} We report the performance of CAT3D in a 3-cond setting evaluated on $512 \times 512$ center-crops.
We use the same datasets and splits as proposed by the original paper.
We measure inference cost in GPU-minutes spent on end-to-end reconstruction process.
In addition, we collect qualitative results shared on the official project websites of ReamDreamer and CAT3D.

\paragraph{Inference Cost.} Our method takes $6.25$ seconds to reconstruct one scene on a single H100 NVIDIA GPU or 15 seconds on an A100. (We report A100 times in~\cref{tab:comparison_optimization}, to be comparable to CAT3D).
CAT3D takes 5 seconds to generate 80 images on 16 GPUs, and needs 640-800 generated images per scene, followed by 4 minutes of reconstruction, amounting to around 5 minutes, depending on the dataset.

\paragraph{Results.}

\begin{figure}
    \centering
    \includegraphics[width=0.99\columnwidth]{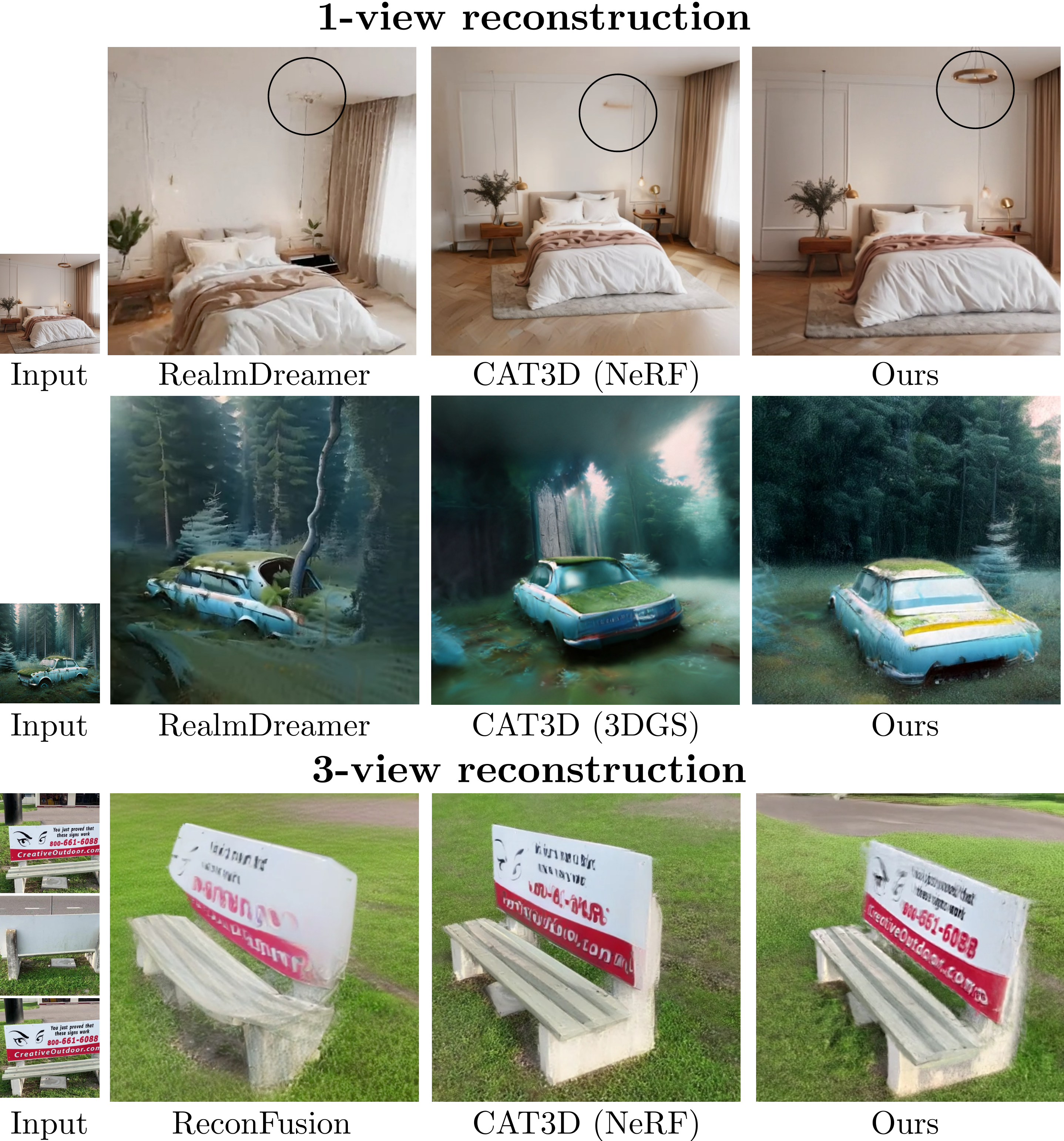}
    \vspace{-6pt}
    \caption{\textbf{Qualitative comparison to optimization-based methods.} The reconstruction quality of our method compares favorably with that of optimization-based methods, and sometimes even exhibits sharper details, while using 300$\times$ less compute.}
    \label{fig:comparison_optimization}
\end{figure}

In~\cref{tab:comparison_optimization} we observe that our method shows strong performance, and it does so while requiring $300\times$ less compute for inference.
Qualitatively, our method produces very high quality reconstructions for a range of scenes, and in~\cref{fig:comparison_optimization} we illustrate that our method generates better results than RealmDreamer, and sometimes sharper results than CAT3D, especially on the 3DGS variant and in backgrounds or fine details.
This is due to the optimization process in CAT3D regressing to the mean in the case of inconsistent generations, especially in the less robust 3DGS optimization process.
While our method does not outperform CAT3D, it is still capable of generating high-quality 3D scenes from a diverse range of inputs (as seen in the supplementary material), and we argue that a $300\times$ reduction in inference cost well justifies a small drop in quality.

\begin{table}
    \centering
    \footnotesize
    \resizebox{0.97\columnwidth}{!}{%
    \begin{tabular}{llccccr}
    \toprule
        & & \multicolumn{4}{c}{\textbf{3-view}} & Infer. Cost \\
        \cmidrule(lr){3-6}  
        & Method  & PSNR $\uparrow$ & SSIM $\uparrow$ & LPIPS $\downarrow$ & FID $\downarrow$ & gpu-min $\downarrow$ \\
        \midrule
        \parbox[t]{2mm}{\multirow{2}{*}{\rotatebox[origin=c]{90}{\scriptsize RE10}}}
        & CAT3D & \textbf{29.56} & \textbf{0.937} & \textbf{0.134} & \textbf{13.75} & 77.28 \\ 
        & Ours & 27.00 & 0.905 & 0.154 & 27.40 & \textbf{0.25} \\
        \midrule
        \parbox[t]{2mm}{\multirow{2}{*}{\rotatebox[origin=c]{90}{\scriptsize LLFF}}}
        & CAT3D & \textbf{22.06} & \textbf{0.745} & \textbf{0.194} & \textbf{37.54} & 80.00 \\
        & Ours & 18.75 & 0.562 & 0.341 & 96.61 & \textbf{0.25} \\
        \midrule
        \parbox[t]{2mm}{\multirow{2}{*}{\rotatebox[origin=c]{90}{\scriptsize DTU}}}
        & CAT3D & \textbf{19.97} & \textbf{0.809} & \textbf{0.202} & \textbf{41.76} & 72.00 \\
        & Ours & 18.59 & 0.738 & 0.312 & 67.49 & \textbf{0.25} \\
        \midrule       
        \parbox[t]{2mm}{\multirow{2}{*}{\rotatebox[origin=c]{90}{\scriptsize CO3D}}}
        & CAT3D & \textbf{20.85} & \textbf{0.673} & \textbf{0.329} & \textbf{44.33} & 73.60 \\
        & Ours & 19.41 & 0.628 & 0.416 & 79.60 & \textbf{0.25} \\
        \midrule       
        \parbox[t]{2mm}{\multirow{2}{*}{\rotatebox[origin=c]{90}{\scriptsize Mip.}}}
        & CAT3D & \textbf{16.62} & \textbf{0.377} & \textbf{0.515} & \textbf{91.82} & 73.60 \\
        & Ours & 15.67 & 0.309 & 0.540 & 124.17 & \textbf{0.25} \\
         \bottomrule
    \end{tabular}
    }
    \caption{\textbf{Comparison to optimization-based methods.} Our method performs competitively with state-of-the-art optimization-based methods, while reducing the inference cost $300\times$. }
    \label{tab:comparison_optimization}
\end{table}

\subsection{Image VAEs generalize poorly to geometry}

\begin{table}[t]
    \centering
    \resizebox{0.97\columnwidth}{!}{%
    \footnotesize
    \begin{tabular}{lccc}
         \toprule
         & \multicolumn{3}{c}{Synth. (Bounded)/Real (Unbounded)} 
         \\
         \cmidrule(lr){2-4}
         & Rel.$\times$100$\downarrow $ & $\delta_{1.01}$$\uparrow$ & $\Delta uv (\text{px})$$\downarrow$  \\ 
        \midrule
        Im AE + mean-z scaling  & 1.03/17.9 & 67.1/24.2 & 9.72/160 \\ 
        Im AE + max-xyz scaling & 4.56/9.74 & 19.8/15.0 & 95.6/907 \\ 
        Im AE + nonlinear scaling & 1.34/15.8 & 58.8/40.3 & 16.9/245 \\
        \midrule
        Geo AE, Conv decoder & \textbf{0.530}/0.684 & \textbf{87.2}/81.4 & 3.94/4.66 \\ 
        Geo AE, ViT decoder \textbf{(Ours)} & 0.636/\textbf{0.670} & 82.4/\textbf{81.5} & \textbf{3.85}/\textbf{2.69}  \\ 
        \bottomrule
    \end{tabular}
    }
    \caption{\textbf{Comparison of VAEs Trained for Images (Im AE) vs. Geometry (Geo AE) on Pointmaps.} Our autoencoder trained specifically for geometry tasks performs much better than pre-trained image auto-encoders, regardless of how points are scaled.
    Our transformer-based decoder outperforms the convolution-based one on real data.}
    \label{tab:geo_decoder}
\end{table}

Typically, VAEs are pre-trained on RGB images for latent diffusion models that generate images~\cite{rombach2022high}. We show that these VAEs do not work well for unbounded geometry.

\paragraph{Metrics.} 
We use metrics commonly used for depth estimation~\cite{ranftl21dpt,wang24dust3r}---relative error (AbsRel) between the target z-component of the pointmap $z$ and prediction $\hat{z}$, Rel. $ = \norm{ z - \hat{z} } / z$, and the prediction threshold accuracy, $ \delta_{\epsilon} = \operatorname{max}(\sfrac{\hat{z}}{z}, \sfrac{z}{\hat{z}}) < \epsilon$ with $\epsilon = 1.01$
Additionally, we measure the re-projection error: the mean Euclidean distance on the image plane when re-projecting the point back to the source camera $\Delta uv = \frac{1}{N}\sum_{i}^N \norm{ (u_i, v_i) - (\hat{u}_i, \hat{v}_i) }_2$, where $u$ and $v$ are the $x$ and $y$ coordinates, respectively, in pixels on the image.
We measure performance separately for synthetic and real data, both at $512\times 512$ resolution.

\paragraph{Scaling.} Our geometry data is not metric, so the scale of the scene is arbitrary.
Properly normalizing the data before inputting it to the VAE can significantly impact performance.
We consider 3 methods for scaling data.
We experiment with (1) scaling the scene such that mean scene depth is $1$, (2) scaling the scene such that the maximum coordinate value in any direction is $1$ and (3) applying a nonlinear contraction function (the sigmoid function) before inputting it to the encoder, and applying its inverse after decoding.
Our autoencoder uses (1).

\begin{figure}[t]
    \centering
     \includegraphics[width=\columnwidth]{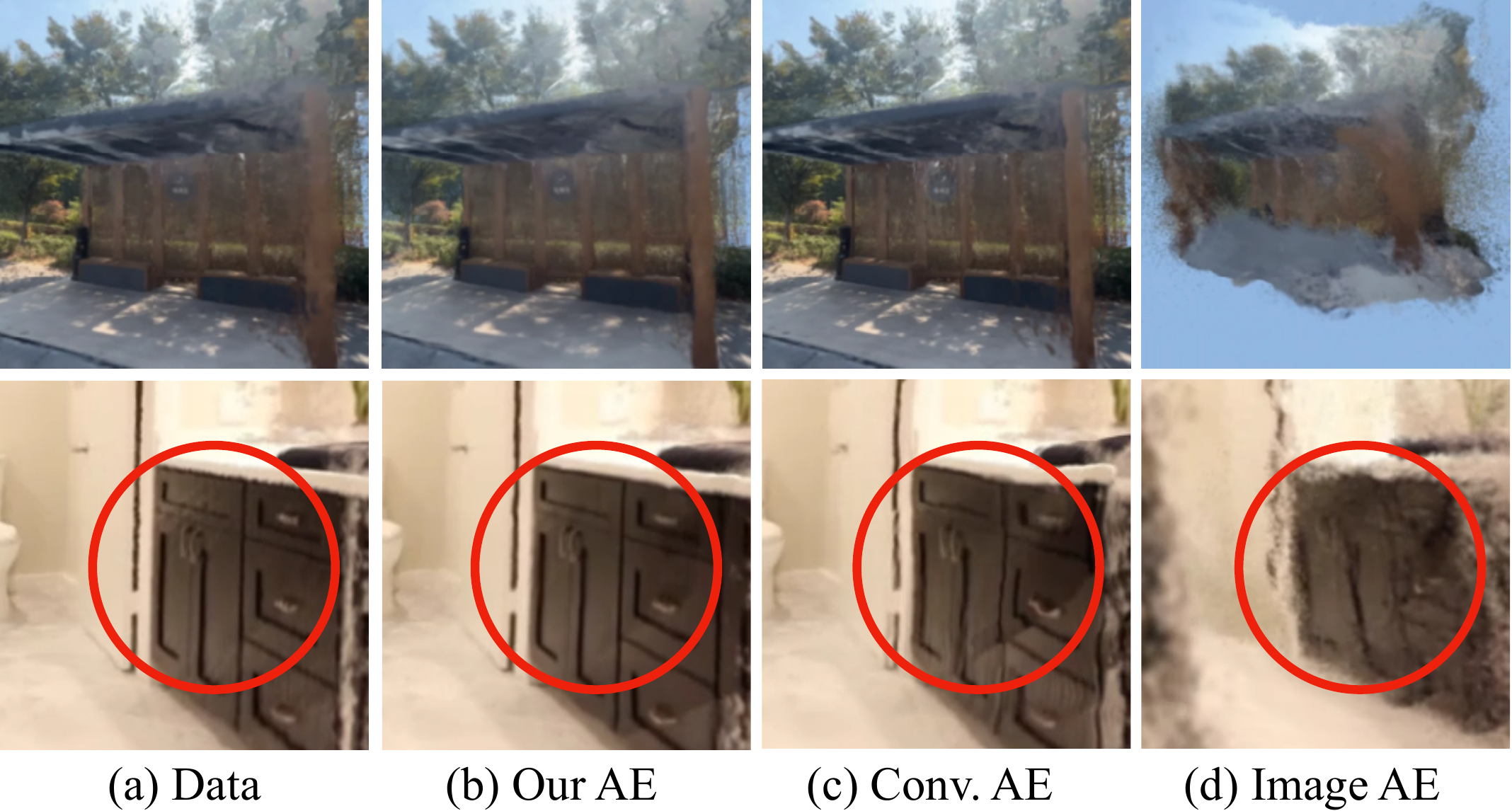}
    \caption{Given input geometry (a), our Auto-Encoder (b) accurately reconstructs input geometry. Using a convolutional decoder (c) results in inaccurate geometry and using a pre-trained Image Auto-Encoder (d) fails catastrophically for both indoor and large-scale outdoor scenes.
    }
    \label{fig:ae_problem}
\end{figure}

\paragraph{Results.} In~\cref{tab:geo_decoder} we find that pre-trained image autoencoders work reasonably well for synthetic, bounded data. This conclusion aligns with prior work showing that VAEs pre-trained on RGB images can be applied effectively to bounded relative depth $d \in [0, 1]$~\cite{marigold}.
However, we find that regardless of the scaling method, autoencoders trained on images struggle to autoencode pointmaps.
Meanwhile, our autoencoder trained on geometry
maintains high accuracy (80$\%$ of points are within $1\%$ of ground truth) for both synthetic and real data.
We visualize 2D pointmaps and point cloud renders in~\cref{fig:ae_problem} and observe that the image autoencoder fails catastrophically in outdoor scenes, and exhibits obvious inaccuracies when autoencoding indoor scenes.
While the convolutional decoder performs comparably to the transformer quantitatively, we empirically observe in~\cref{fig:ae_problem} that it introduces jarring artifacts, e.g., bent lines.
This analysis illustrates that our Geometry AE and its architecture are instrumental to high-quality results on unbounded scenes from our method. 

\section{Conclusion}%
\label{sec:conclusion}

We have presented \method, a fast feed-forward method that generates detailed 3D scenes in less than 7 seconds. To enable this capability, we propose a 3D scene representation that allows for denoising high-resolution 3D scenes using powerful 2D latent diffusion architectures. We create a large-scale 3D scene dataset to train \method, and demonstrate that it accurately models ambiguity, enabling high-quality 1-view reconstructions where regression-based methods fail.
Our proposed feed-forward approach is capable of reconstructing a wide variety of scenes and reduces the cost of 3D generation by $300\times$ compared to existing optimization-based methods, opening up opportunities for 3D content creation at scale.

\paragraph{Acknowledgments}
We would like to express our deepest gratitude to Ben Poole for helpful suggestions, guidance, and contributions. We also thank George Kopanas, Sander Dieleman, Matthew Burruss, Matthew Levine, Daniel Duckworth, Peter Hedman, Songyou Peng, Rundi Wu, Alex Trevithick, Hadi Alzayer, David Charatan, Jiapeng Tang and Akshay Krishnan for valuable discussions and insights. Finally, we extend our gratitude to Shlomi Fruchter, Kevin Murphy, Mohammad Babaeizadeh, Han Zhang and Amir Hertz for training the base text-to-image latent diffusion model. This work was done while S. Szymanowicz was  on an internship at Google. He is supported by the EPSRC Doctoral Training Partnerships Scholarship (DTP)
EP/R513295/1 and the Oxford-Ashton Scholarship.

{
    \small
    \bibliographystyle{ieeenat_fullname}
    \bibliography{main}
}

\clearpage
\maketitlesupplementary

\section{More experimental results}

\subsection{Videos and interactive results}

The project website contains interactive results and video results, which we encourage the reader to explore.

\subsection{Comparison to video models}

\begin{figure}[h]
    \centering
    \includegraphics[width=0.9\columnwidth]{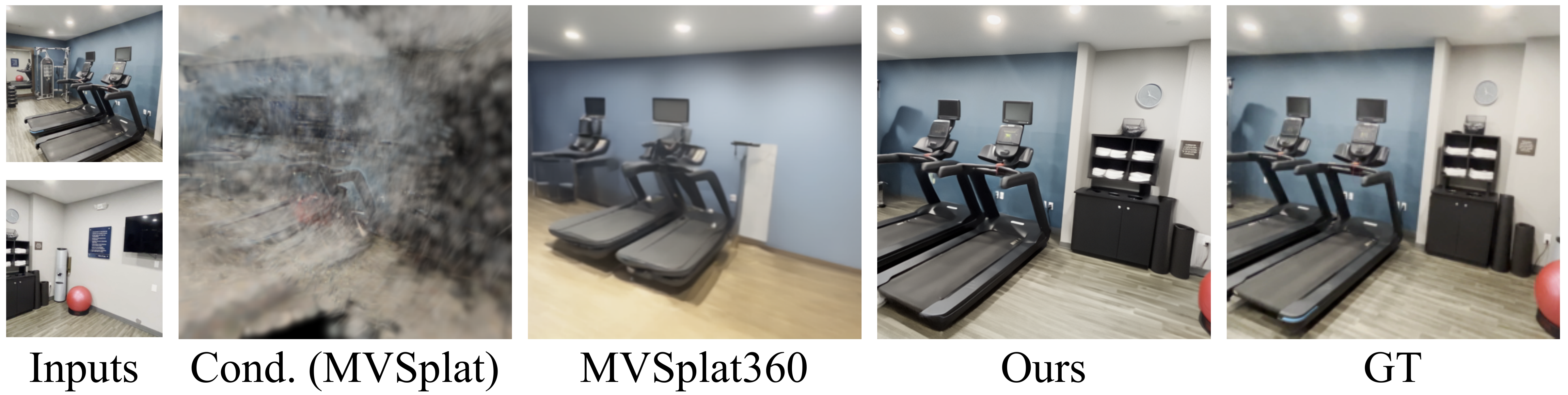}
    \caption{MVSplat360 uses MVSplat for conditioning, which works poorly when there is little or no input view overlap.}
    \label{fig:mvsplat360}
\end{figure}

Video models have emerged as a powerful tool for novel view systhesis. 
However, using them to reconstruct a 3D asset from a generated video requires distillation~\cite{liu2024reconx}, similarly to CAT3D, which significantly increases the runtime of such approaches when applied to 3D reconstruction.
Nonetheless, we evaluate the quality of novel views generated by one representative approach, MVSplat360~\cite{chen2025mvsplat360}.
We evaluate Bolt3D and MVSplat360 on 2-view and 4-view reconstruction on scenes from DL3DV at $512\times512$ resolution.
We take care to input appropriately-sized renders to MVSplat360's diffusion model to match its training resolution, and crop images appropriately for evaluation.
In~\cref{tab:mvsplat360} we observe that renders from Bolt3D's 3D scenes are more accurate than the novel views generated by MVSplat360.
In~\cref{fig:mvsplat360} we find that the conditioning often needed for video models (e.g., ViewCrafter~\cite{yu2024viewcrafter}, MVSplat360, ReconX~\cite{liu2024reconx}) can be brittle, resulting in poor accuracy of generated views.
In addition, video models are also slow (5.8 minutes for 56 frames) while Bolt3D reconstructs a 3D asset in 7 {\bf seconds} and renders in real-time.

\begin{table}[h]
    \centering
    \footnotesize
    \resizebox{0.98\columnwidth}{!}{%
    \begin{tabular}{llcccccc}
    \toprule
        {} & Method & PSNR $\uparrow$ & SSIM $\uparrow$ & LPIPS $\downarrow$ & FID $\downarrow$ \\
        \midrule
        2-view & MVSplat360 & 13.97 & 0.400 & 0.575 & 104.83 \\
        DL3DV & Ours & \textbf{17.75} & \textbf{0.550} & \textbf{0.392} & \textbf{64.53} \\
        \midrule
        4-view & MVSplat360 & 15.42 & 0.422 & 0.507 & 76.91 \\
        DL3DV & Ours & \textbf{20.64} & \textbf{0.652} & \textbf{0.311} & \textbf{48.28} \\
        \bottomrule
    \end{tabular}
    }
    \caption{\textbf{MVSplat360}'s video model renderer is less accurate and more than 200$\times$ slower than Bolt3D for 2- and 4-view DL3DV.}
    \label{tab:mvsplat360}
\end{table}

\subsection{Ablations}

\paragraph{Ablation---Geometry VAE.}
We observe in~\cref{tab:geo_vae_ablation} that training the encoder (rather than using a frozen, pre-trained one) is important for high autoencoding precision, likely due to pointmaps being outside of the value range on which the encoder was pre-trained.
Removing the weighting on distant points (\cref{eq:mipnerf_supplement}) or the gradient loss (Eq. 7 main paper) reduces performance of our system.

\paragraph{Ablation---Gaussian Head.} 
In~\cref{tab:appearance_ablation} we illustrate that the design choices in the Gaussian Head are important for high-quality rendering.
Using fewer (8, rather than 16) views reduces scene coverage and thus incurs a bigger rendering error.
Cross attention is important because it allows modulating opacity in splatter images depending on visibility from other views.
Using constant opacity and scale parameters reduces rendering quality.
Interestingly, allowing the means to be modified by the Gaussian Head also drops performance, suggesting that explicit geometry losses give a stronger supervisory signal, consistent with comparisons to Wonderland~\cite{liang2024wonderland} in the main paper.
Lastly, forcing the Gaussians to lie on rays is advantageous for rendering quality.

\paragraph{Low-resolution comparisons}

Most methods evaluate performance at lower resolution than our method can handle, and sometimes train at a different resolution, making apples-to-apples comparison challenging.
We make the best effort at comprehensive testing at different possible input resolutions to present baseline methods in the most favorable chance.
First, we verify that the best way to evaluate baseline methods at high resolution is to feed in a high resolution image~\cref{tab:regressive_reconstruction_upres}, rather than upscaling outputs from lower-resolution input.
This is the method we use for all baseline methods in the main paper.

Next, we evaluate performance at a lower resolution, closer to the training setup of baseline methods.
In~\cref{tab:regressive_reconstruction_lowres} we illustrate that at a lower $256\times256$ resolution, our model still outperforms Depthsplat.
Finally, we give Depthsplat an advantage by evaluating its performance when receiving wide field-of-view (fov) $256\times448$ images.
In this setting, the wider field-of-view provides more scene coverage by a factor of $\times 1.75$ and more cues for matching across different views.
Only in this setting do we find that Depthsplat achieves similar performance to our method.

\begin{table}[]
    \centering
    \footnotesize
    \begin{tabular}{l c c c}
        \toprule
        {} &  Rel. $\downarrow$ & $\delta_{1.01}$ $\uparrow$ & $\Delta uv (\text{px}) \downarrow$ \\
        \midrule
        Full model & \textbf{0.99} & \textbf{73.0} & \textbf{2.56} \\
        \midrule
        - encoder training & 1.63 & 58.9 & 3.79 \\
        - $\mathcal{L}_\text{rec}$ re-weighting & 1.68 & 56.7 & 5.43 \\
        - $\mathcal{L}_\text{grad}$ & 1.16 & 69.8 & 2.96 \\
        \bottomrule
    \end{tabular}
    \caption{\textbf{Geometry ablation.} Removing encoder training or geometry-specific losses leads to worse performance.}
    \label{tab:geo_vae_ablation}
\end{table}

\begin{table}[]
    \centering
    \footnotesize
    \begin{tabular}{l c c c}
        \toprule
              & PSNR $\uparrow$ & SSIM $\uparrow$ & LPIPS $\downarrow$ \\
        \midrule
        Ours & \textbf{24.72} & \textbf{0.831} & \textbf{0.209} \\
        \midrule
        Fewer views at inference & 24.30 & 0.823 & 0.221 \\
        No cross-attention & 23.80 & 0.804 & 0.239 \\
        No Gaussian Head & 21.94 & 0.734 &  0.343 \\
        XYZ learnt from rendering & 21.88 & 0.755 & 0.311 \\
        No ray-clipping & 20.78 & 0.642 & 0.288 \\
        \bottomrule
    \end{tabular}
    \caption{\textbf{Appearance ablation.}
    All components of architecture, training and inference are important for high-quality appearance.}
    \label{tab:appearance_ablation}
\end{table}

\begin{table}
    \centering
    \footnotesize
    \begin{tabular}{l ccc}
    \toprule
         Resolution input   \\
         to network & PSNR $\uparrow$ & SSIM $\uparrow$ & LPIPS $\downarrow$ \\
        \midrule
         256 $\times$ 256 & 23.49 & 0.845 & 0.203 \\
         512 $\times$ 512 & \textbf{24.17} & \textbf{0.875} & \textbf{0.183} \\
         \bottomrule
    \end{tabular}
    \caption{\textbf{Evaluating Depthsplat at high resolution.} 
    We verify that the most advantageous setting for Depthsplat when evaluating it on high $512\times 512$ resolution is to use a high-resolution input.
    }
    \label{tab:regressive_reconstruction_upres}
\end{table}

\begin{table}
    \centering
    \scriptsize
    \begin{tabular}{clcccc}
    \toprule
         & Method & Input Res.   & PSNR $\uparrow$ & SSIM $\uparrow$ & LPIPS $\downarrow$ \\
        \midrule
        \multirow{2}{1cm}{1-view RE10K}        & Flash3D  & 256$\times$256 & 17.70 & 0.616 & 0.393 \\
                            & Ours & 256$\times$256 & \textbf{21.62} & \textbf{0.804} & \textbf{0.202} \\
        \midrule
         \multirow{2}{1cm}{3-view RE10K} & Depthsplat  & 256$\times$256 &  24.69 & 0.873 & 0.126 \\
                     & Ours & 256$\times$256 & \textbf{27.39} & \textbf{0.916} & \textbf{0.103} \\
        \midrule
        \multirow{3}{1cm}{2-view DL3DV} & Depthsplat  & 256$\times$448 & 18.09 & 0.549 & 0.323 \\\cmidrule(lr){2-6}
        & Depthsplat  & 256$\times$256  & 16.16 & 0.467 & 0.388  \\
                     & Ours & 256$\times$256 & \textbf{18.01} & \textbf{0.556} & \textbf{0.320} \\

        \midrule
        \multirow{3}{1cm}{4-view DL3DV} & Depthsplat  & 256$\times$448 & 21.20 & 0.697 & 0.208 \\\cmidrule(lr){2-6}
        & Depthsplat  & 256$\times$256 & 19.64 & 0.633 & 0.254 \\
                     & Ours & 256$\times$256 & \textbf{21.16} &\textbf{ 0.695} & \textbf{0.231} \\

        \midrule
        \multirow{3}{1cm}{6-view DL3DV} & Depthsplat  & 256$\times$448 & 21.93 & 0.730 & 0.184 \\
        \cmidrule(lr){2-6}
        & Depthsplat  & 256$\times$256 & 20.64 & 0.680 & 0.225 \\
                     & Ours & 256$\times$256 & \textbf{22.18} & \textbf{0.733} & \textbf{0.206} \\

         \bottomrule
    \end{tabular}
    \caption{\textbf{Low-resolution 256$\times$256 comparisons.} Our method outperforms competitors at low resolution when receiving the same input information.
    Only when DepthSplat~\cite{xu2025depthsplat} receives 1.75 times more input information than our method by ingesting wide-fov images, does its performance become similar to ours.
    }
    \label{tab:regressive_reconstruction_lowres}
\end{table}

\section{Implementation details}

\subsection{XYZ normalization}
\label{s:encoding}

\paragraph{Relativization.} 
The supervising (pseudo-ground truth) data used to train our diffusion model is reconstructed using off-the-shelf 3D reconstruction algorithms (MASt3R)~\cite{mast3r}. We transform this 3D reconstruction to the view-space of the first camera, such that all point coordinates and all cameras are relative to this coordinate frame: $\mat{\Pi}, \mat{\I}^{\superscript{xyz}} := \mat{\Pi}\mat{\Pi}_0^{-1}, \mat{\I}^{\superscript{xyz}} \mat{\Pi}_{0}^{-1}$, where $\mat{\Pi}_{0}$ denotes the camera-to-world rigid body transform of the first camera.

\paragraph{Scaling.}
We normalize the 3D scale of the reconstructed scenes by applying a per-scene scaling factor $\alpha$ to the camera poses and point coordinates: $\mat{\Pi}, \mat{\I}^{\superscript{xyz}} = \alpha \mat{\Pi}, \alpha \mat{\I}^{\superscript{xyz}}$. This scale factor is chosen such that the mean depth value from the first camera is the same across every scene in our dataset:
$ \alpha \bar{\mat{\I}}^{\superscript{xyz}}_{0}[z] = 1 $.

\paragraph{Re-weighting Points in VAE Reconstruction Loss.} In our VAE reconstruction loss (Eq 5. main paper), we introduce a re-weighting scheme for two reasons: 1) ground truth points far from the scene center are more likely to be incorrect, and 2) points with high magnitude would make up a large proportion of an equal weighting loss.

For each scene, the point maps are defined in the coordinate system of the ``first camera." When computing the reconstruction loss, we first transform each point $\bx\in \mat{P}$ to the local camera coordinate system:
\begin{equation}
    \bx_\text{local} = \left[R ~ | ~ T\right]_\text{w2c} \bx,
\end{equation}
where $\left[R ~ | ~ T\right]_\text{w2c}$ is the world-to-camera transformation matrix.
Because scenes are scaled such that the mean depth to the first camera is 1, we can think of $[0, 0, 1]^\top$ as the look-at point or center of the scene, so $d = \norm{\bx_\text{local} - [0, 0, 1]^\top}$ is the distance of the point to the center of the scene of the local camera. Thus, we compute:
\begin{equation}\label{eq:mipnerf_supplement}
    \mathcal{L}_{\text{rec}} = \frac{2\sqrt{w} - 1 }{w} \norm{\hat{\bx}_\text{local} - \bx_\text{local, gt}}^{2}
\end{equation}
where $w = \max(1, d^2)$ is the bounded squared distance to the local scene center and $\frac{2\sqrt{w} - 1 }{w}$ is the Jacobian of the contraction function defined in MipNerf-360~\cite{barron2022mipnerf360}.



\subsection{Architecture and training details.}\label{s:multiview_denoising}

\paragraph{Diffusion model.}
We use a U-Net with full 3D attention on all feature maps up to $32\times32$, as in CAT3D~\cite{gao2024cat3d}.
Unlike CAT3D, our diffusion model is trained to model the joint distribution of latent appearance and geometry.
To this end, we increase the number of channels in the input and output layers of CAT3D's architecture by 8 to additionally accept geometry latents.
The input to our network thus has 8-dimensions for the geometry latents, 8-dimensions for the image latent, 6-dimensions for the camera pose raymaps, and a 1-dimensional mask indicating which views are given as conditioning, yielding an input dimension of $64\!\times\!64\!\times\!23$.
We train with the same optimization hyperparameters as~\cite{gao2024cat3d}, except we additionally finetune on $16$ input views with a lower learning rate of $1e-5$.

\paragraph{Autoencoder.}
We use a pre-trained and frozen image autoencoder similar to that of Stable Diffusion~\cite{rombach2022high}.
The geometry encoder has the same architecture, except we increase the channel dimension to additionally accept a 6-dimensional camera pose representation.
The decoder is a transformer-based network.
We patchify the $64\times64\times8$ latent with patch size $2$, thus using a token length $1024$.
We use the ViT-B architecture hyperparameters: 12 layers with channel size $768$, with the fully-connected layer consisting of 2 dense layers with GeLU activation function and a hidden MLP dimension $3072$.
The linear projection head projects each token to a $16\times16$ patch.
We optimize the parameters of the Autoencoder with the Adam~\cite{kingma15adam} optimizer using constant learning rate $1e-4$, batch size $512$, Adam parameters $(\beta_1, \beta_2) = (0.0, 0.99)$.
We first train for 3M iterations at $256\times256$ resolution, followed by fine-tuning at $512\times512$ for 250k iterations.
We use loss weight parameters $\lambda_1=3e-9$ and $\lambda_2=0.033$.

\paragraph{Foreground masking for synthetic data.}

In synthetic data, we apply loss $\mathcal{L}_{rec}$ only on foreground pixels and train the model to additionally output a foreground alpha mask, supervised with binary cross-entropy loss.

\begin{figure*}
    \centering
    \includegraphics[width=6in]{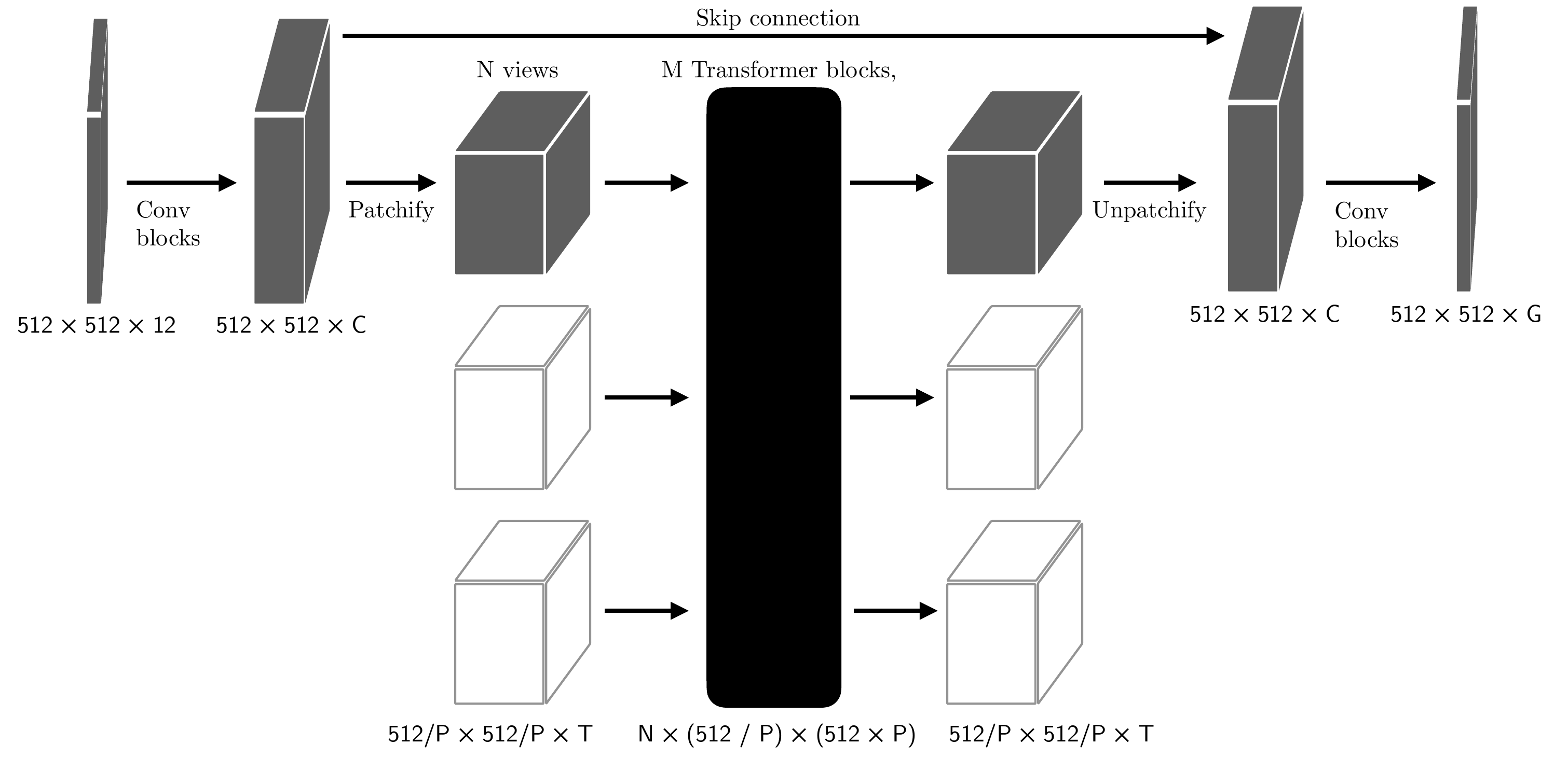}
    \caption{Our Gaussian head architecture consists of convolutional and transformer blocks, with patchification to manage the cross-attention sequence length.}
    \label{fig:gaussian_head_arch}
\end{figure*}

\paragraph{Gaussian head.}
We detail the Gaussian head architecture in~\cref{fig:gaussian_head_arch}.
The Gaussian Head receives as input the 3-dimensional image, 3-dimensional pointmap, and 6-dimensional raymap encoding the camera pose.
Each view is first passed through 3 residual convolutional blocks with the Swish activation function~\cite{ramachandran2017searching} and channel size $128$, followed by $4\times4$ patchification to token dimension $128$ and full cross attention. We use 3 transformer layers with hidden dimension $128$ and $8$ heads, MLP dimension $512$.
The tokens are then unpatchified to original resolution and $128$ channel dimension.
Following that, there are another 3 residual convolutional blocks, and a final, unactivated $3\times3$ convolutional layer that outputs channel size $11$ ($3$ for color, $3$ for size, $4$ for rotation and $1$ for opacity).
The outputs are activated with an exponential function for scale and a sigmoid function for opacity.
To facilitate accurate scale prediction, the size output by the network is then multiplied by the z-distance of the gaussian from the camera.
The means of Gaussians are not predicted by the Gaussian head, as they are already available from the VAE.
The Gaussian head is trained with $8$ input views, leading to a sequence length $131k$.
We manage this computational complexity by using FlashAttention~\cite{dao2022flashattention,dao2023flashattention2} and rematerializing gradients on the dot product operation.
For losses, we use L2 photometric loss with weight $\lambda=1$ and LPIPS loss weight $\lambda_{LPIPS} = 0.05$.
We train with learning rate $1e-4$ and batch size $8$.
When training the Gaussian Head, the Geometry and Image autoencoders are frozen.

\paragraph{Gaussian head for viewer assets}

To enhance rendering performance and reduce the memory footprint of assets for the viewer on our project website, we add an L1 regularizer term to encourage completely transparent Gaussians when they are not necessary, similar to LongLRM~\cite{ziwen2024longlrm}.
Gaussians with low opacity are then culled before saving the asset.
To further reduce file sizes, the model data is quantized in chunks of 256 gaussians (\url{https://github.com/playcanvas/splat-transform}).

\subsection{Inference.}

\paragraph{Sampling details.}
We train our diffusion model~\cite{ho2020denoising} using $v$-parameterization~\cite{salimans2022progressive} with $T$=1000 timesteps. At inference time, we use DDIM~\cite{song2020denoising} to speed up inference to 50 steps using the same noise schedule as CAT3D~\cite{gao2024cat3d} except with zero terminal SNR~\cite{lin2023common}.

\paragraph{Camera path sampling.}
We use the same camera path heuristics as CAT3D -- sampling circular paths, forward-facing paths and splines.
We use much fewer views than CAT3D (16 vs their 800), so we sample camera paths on only one path, typically with the median radius and height of cameras in the training set, without offsets or scaling~\cite{gao2024cat3d}.

\section{Limitations, discussion and future work}

\paragraph{Limitations.} While our method can produce a wide range of geometries, it still struggles on thin structures, especially those that are fewer than 8 pixels wide (the spatial downsampling ratio of our geometry VAE).
Our method also struggles with scenes that have large amounts of transparent or highly non-lambertian surfaces, for which geometry reconstruction in Structure-from-Motion frameworks is typically inaccurate.

Our model is also sensitive to the distribution of the target cameras, in particular the up-vector chosen to generate the camera path as well as the scene scale. Perhaps these could ameliorated in future work with better data augmentation. 

\paragraph{Discussion and future work.} To the best of our knowledge, Bolt3D is the first work to explore the architecture and training recipe of a Geometry VAE, and there remain several design choices to be explored.
In particular, we chose compress pointmaps over depth due to their resounding success in multi-view reconstruction (Dust3r~\cite{wang24dust3r}, Mast3r~\cite{mast3r}), but concurrent work shows that inferring depth can be complimentary~\cite{jiang2025geo4d} or even advantageous~\cite{wang2025vggt}.
We leave exploration of these findings in context of 3D generation to future works.

Next, despite making a significant step in feed-forward 3D generation, the quality of 3D generation, Bolt3D lacks in quality compared to optimization-based methods such as CAT3D.
We hypothesize this is due to CAT3D generating much more views ($\approx$800, compared to our 16), resulting in more complete scene coverage.
Generating more views, perhaps through an anchoring strategy~\cite{sargent2023zeronvs}, could improve the quality of results, though it would result in a large number of 3D Gaussians.

Finally, Bolt3D generates exclusively static scenes.
Perhaps future work could combine multi-view video diffusion models~\cite{wu2025cat4d} with Bolt3D's direct geometry generation to generate dynamic 3D scenes in a feed-forward manner.

\section{Experimental details}
\paragraph{DL3DV scenes.}

We ran evaluation on the intersection of our test set and the public test benchmark. 
In 2-view, 4-view and 6-view setups we used center-crops of views $(0, 49)$, $(0, 19, 29, 49)$ and $(0, 9, 19, 29, 39, 49)$, respectively.  
The scenes used for evaluation were: 
\begin{itemize}
\item \hash{0569e83fdc248a51fc0ab082ce5e2baff15755c53c207f545e6d02d91f01d166},
\item \hash{073f5a9b983ced6fb28b23051260558b165f328a16b2d33fe20585b7ee4ad561},
\item \hash{183dd248f6a86e07c5adf9de8ee2d0abe45b1216331c03678e89634c2e9b1c7f},
\item \hash{1ba74c22670ad047981441581d00f26f4a148d1010bcac7468c615adf5fa4d5d},
\item \hash{389a460ca1995e0658e85fe8e6b520b4e88b370cd6710dfe728b1564bba31aee},
\item \hash{493816813d2d6d248eb3c2b0b77b63e54235266e9a06e270fd0d282f13960493},
\item \hash{50c46cf8b8b22c8d2ffdef8964b05ddbceaef312c9a9ff331d1ecebfd223f72a},
\item \hash{4ae797d07b6d1644c9db6919c8cc8c0d28d72be45108ac7a3abf8dc21b599d83},
\item \hash{565553aa894be621e8b4773cac288e60ad0c2cf7edb621be62b348c9a0f78380},
\item \hash{599ca3e04cae3ec83affc426af7d0d7ab36eb91cd8e539edbc13070a4d455792},
\item \hash{5c8dafad7d782c76ffad8c14e9e1244ce2b83aa12324c54a3cc10176964acf04},
\item \hash{63798f5c6fbfcb4eb686268248b8ecbc8d87d920b2bcce967eeaedfd3b3b6d82},
\item \hash{946f49be73928469000baa5ca04d2573137c5ee6a66362bcf8d130354dca8924},
\item \hash{9e9a89ae6fed06d6e2f4749b4b0059f35ca97f848cedc4a14345999e746f7884},
\item \hash{cd9c981eeb4a9091547af19181b382698e9d9eee0a838c7c9783a8a268af6aee},
\item \hash{d4fbeba0168af8fddb2fc695881787aedcd62f477c7dcec9ebca7b8594bbd95b}.
\end{itemize}

\end{document}